\documentclass{article}
\usepackage{iclr2027_conference,times}
\usepackage[T1]{fontenc}
\usepackage{amsmath,amssymb}
\usepackage{graphicx,booktabs,array}
\usepackage{placeins,needspace}
\usepackage{colortbl}
\usepackage{microtype}
\definecolor{CaseThink}{rgb}{0.32,0.20,0.85}
\definecolor{CaseAction}{rgb}{0.00,0.50,0.70}
\definecolor{CaseInfo}{rgb}{0.75,0.36,0.10}
\definecolor{CaseAnswer}{rgb}{0.80,0.12,0.36}
\definecolor{CaseShade}{gray}{0.95}
\definecolor{PromptHeader}{gray}{0.90}
\newcommand{\CaseTag}[2]{\textcolor{#1}{\texttt{<#2>}}}
\newcommand{\PromptField}[1]{\textcolor{CaseAction}{\texttt{\{#1\}}}}
\newenvironment{PromptPanel}[1]{%
  \par\noindent\begin{minipage}{\linewidth}
  \fontsize{9}{10.5}\selectfont
  \setlength{\parskip}{2pt}%
  \setlength{\tabcolsep}{5pt}%
  \renewcommand{\arraystretch}{1.08}%
  \begin{tabular}{>{\raggedright\arraybackslash}p{\dimexpr\linewidth-2\tabcolsep\relax}}
  \hline\hline
  \rowcolor{PromptHeader}\textbf{#1} \\
  \hline
}{%
  \\ \hline\hline
  \end{tabular}\end{minipage}\par
}
\usepackage{hyperref}
\usepackage{url}
\hypersetup{colorlinks=true,citecolor=blue,linkcolor=blue,urlcolor=blue,
  pdftitle={UniOPSD: Unifying Outcome and Hindsight Feedback for Agentic Reinforcement Learning},
  pdfauthor={Zenghuang Fu, Zhaoyang Li, Qiuyuan Ai, Xiaofeng Han, Zelong Zheng, Haoyu Wu, Tianyu Fu, Chenxu Zhao, Minghui Wu, Guannan He, Changwei Wang}}

\title{UniOPSD: Unifying Outcome and Hindsight Feedback for Agentic Reinforcement Learning}
\author{%
\begin{minipage}[t]{\dimexpr\textwidth-2\tabcolsep\relax}
\centering\normalsize\bfseries
Zenghuang Fu\textsuperscript{1,2,*}\quad
Zhaoyang Li\textsuperscript{3,*}\quad
Qiuyuan Ai\textsuperscript{3,*}\\[3pt]
Xiaofeng Han\textsuperscript{1,2}\quad
Zelong Zheng\textsuperscript{1,2}\quad
Haoyu Wu\textsuperscript{4}\quad
Tianyu Fu\textsuperscript{4}\\[3pt]
Chenxu Zhao\textsuperscript{4}\quad
Minghui Wu\textsuperscript{4}\quad
Guannan He\textsuperscript{3,\textdagger}\quad
Changwei Wang\textsuperscript{5,6,\textdagger}\\[7pt]
\normalfont\small
\textsuperscript{1}University of Chinese Academy of Sciences\\
\textsuperscript{2}Institute of Automation, Chinese Academy of Sciences\\
\textsuperscript{3}Peking University\quad
\textsuperscript{4}Mininglamp Technology\\
\textsuperscript{5}Key Laboratory of Computing Power Network and Information Security,\\
Ministry of Education; Shandong Computer Science Center,\\
Qilu University of Technology (Shandong Academy of Sciences)\\
\textsuperscript{6}Key Laboratory of Computing Power Internet and Service Computing,\\
Shandong Fundamental Research Center for Computer Science\\[5pt]
\footnotesize
\textsuperscript{*}Equal contribution.\quad
\textsuperscript{\textdagger}Corresponding authors.
\end{minipage}}
\iclrfinalcopy

\begin{document}
\maketitle
% The template's final mode sets a conference publication header; omit it for arXiv.
\lhead{}
% ===== ABSTRACT =====
\begin{abstract}
Reinforcement learning has become an effective approach to training language model agents, but sparse and delayed outcome rewards provide limited guidance for credit assignment across long interaction sequences. Recent work on on-policy self-distillation (OPSD) offers complementary supervision by evaluating a policy's sampled responses under privileged training-time context. However, our diagnostics show that positive average agreement between outcome and hindsight feedback coexists with substantial local disagreement, raising the question of how to allocate influence between them at each decision. We introduce UniOPSD (Unified On-Policy Self-Distillation), which unifies these feedback sources through adaptive local credit arbitration. UniOPSD constructs comparable credit estimates from environmental returns and successful-peer hindsight at shared interaction anchors. Historical agreement determines the global mixing level, while current signal availability and relative precision adjust each source's influence at individual decisions. The episode-level outcome contribution is retained, and bounded token modulation refines the fused step credit for policy optimization. With Qwen2.5-3B-Instruct and Qwen2.5-7B-Instruct, UniOPSD achieves ALFWorld success rates of $82.8\%$ and $83.6\%$, WebShop success rates of $75.0\%$ and $82.0\%$, and Search-QA aggregate accuracies of $45.3\%$ and $49.8\%$, respectively. On 3B WebShop, UniOPSD improves over SDAR by $7.0$ percentage points.
Our code is available at \url{https://github.com/Zenghuang-Fu/Uniopsd}.
\end{abstract}
% ===== INTRODUCTION =====
\section{Introduction}
\label{sec:introduction}

Interactive language agents must coordinate multiple decisions before receiving a task outcome, whether manipulating a household environment, selecting a product, or searching for evidence~\citep{alfworld2021,webshop2022,searchr12025}. A failed trajectory may contain useful intermediate actions, while a successful one may include unnecessary steps. Assigning the same outcome signal to every decision obscures these differences. The challenge is to extract informative local supervision from complete interactions without assuming that every available signal measures the same aspect of action quality.

Outcome feedback and hindsight provide complementary information for learning interactive language-model policies. Outcome feedback relates behavior to task returns: group-relative policy optimization compares sampled trajectories, and GiGPO additionally compares decisions at repeated interaction anchors~\citep{grpo2024,gigpo2025}. These local comparisons can be limited by singleton anchors or identical observed returns, leaving little information with which to distinguish actions. Hindsight can supply another perspective by allowing the policy to reconsider a sampled response in light of a successful trajectory. Existing self-distillation methods already exploit privileged feedback: SDAR and StepOPSD use privileged contexts for agent learning, while AgentOPSD constructs turn-level hindsight evidence~\citep{sdar2026,stepopsd2026,agentopsd2026}. Such feedback can expose useful action patterns that a small set of outcome comparisons fails to distinguish. The potential complementarity motivates using both sources, with environmental returns grounding the update and hindsight supplying additional information about local decisions.

Complementary supervision, however, does not imply consistent local credit. At the same anchor, a sampled action may receive positive hindsight-derived credit and negative outcome-derived credit, because the two assessments rely on different evidence. Hindsight support can reflect imitation or wording preferences as well as useful behavior~\citep{feedbackalign2026,sdsddiversity2026}, while outcome credit depends on the continuations actually sampled. Fixed mixing cannot adapt its balance to this changing evidence, and weighting by local stability alone can favor a consistent but return-irrelevant teacher. Existing work addresses parts of this problem: StepOPSD shapes sampled-action advantages with hindsight gaps, and ADRS relates teacher confidence to returns, including within GiGPO anchors, to construct token modulation~\citep{stepopsd2026,adrs2026}. We focus on the allocation of influence between two explicit local credit estimates before token modulation.

Our training diagnostics show why average agreement is insufficient. Over updates 1--150, ALFWorld-3B has mean within-batch correlation $0.315$, but mean sign disagreement reaches $37.8\%$ among rows carrying both signals; joint coverage averages only $32.8\%$ (Appendix~\ref{app:diagnostics}). Positive association thus coexists with local disagreement and missing evidence. These batch-level summaries motivate separating historical agreement from the availability and relative precision of current evidence. This raises the central question: \emph{when outcome and hindsight credit disagree, which signal should an agent trust, and by how much?}

We introduce UniOPSD (Unified On-Policy Self-Distillation), which unifies outcome and hindsight feedback by adaptively arbitrating between their local credit estimates. The two estimates are GiGPO's weighted, mean-centered return-to-go and a hindsight branch obtained by centering and scale-matching the frozen behavior policy's log-probability gaps at the same anchors. Historical agreement provides a reliability proxy that sets the global mixing level, while current signal availability and relative precision adjust each step's weight. Convex fusion expresses the resulting allocation of influence. The episode-level outcome term remains numerically fixed, and bounded token modulation refines the arbitrated step credit before PPO, as illustrated in Figure~\ref{fig:method}. We evaluate UniOPSD on ALFWorld, WebShop, and Search-QA with Qwen2.5-3B-Instruct and Qwen2.5-7B-Instruct. The benchmark results show competitive performance across these domains, including a $7.0$-percentage-point improvement in 3B WebShop success over the strongest reported baseline in Table~\ref{tab:reported-baselines}. Credit-signal analysis reveals substantial local disagreement despite positive average association, connecting the empirical setting to the arbitration problem.

Our contributions are threefold:
\begin{itemize}
\item \textbf{A formulation of local credit arbitration.} We express outcome and hindsight supervision as comparable step-credit branches at matched anchors, making disagreement and unequal evidence availability explicit while retaining the native episode term.
\item \textbf{Arbitration using historical and local evidence.} We develop the \texttt{corr} rule, which uses historical agreement to set a global mixing level and local precision proxies and availability to adjust each step's weight, followed by bounded token refinement.
\item \textbf{Empirical evaluation and analysis.} We report benchmark results across three agent domains and two model scales, together with component ablations and diagnostics of local disagreement, evidence coverage, and learning trajectories. A token-perturbation bound and exact reductions separate the roles of step arbitration and token refinement.
\end{itemize}

% ===== RELATED WORK =====
\section{Related Work}
\label{sec:related}

\paragraph{Agent reinforcement learning.}
Policy-gradient methods provide trajectory- and step-level learning signals~\citep{gae2015,ppo2017,grpo2024,gigpo2025}. Hindsight, rollout graphs, and pivot supervision refine credit assignment~\citep{hca2019,hcapo2026,graphgpo2026,pica2026}, while ActFocus reweights action tokens~\citep{actfocus2026}. PCGrad and CAGrad address task-gradient conflicts~\citep{pcgrad2020,cagrad2021}; UniOPSD instead allocates outcome and hindsight credit for individual decisions.

\paragraph{On-policy self-distillation.}
Distillation supplies teacher supervision, including on student-generated responses~\citep{distillation2015,gkd2024}; privileged contexts and temporal curricula extend this approach to reasoning and interaction~\citep{opsd2026,tcod2026,sdar2026,stepopsd2026}. RLSD, SDPG, and PBSD connect distillation to advantage reweighting, verifier-based updates, and preference optimization, respectively~\citep{rlsd2026,sdpg2026,pbsd2026}. Feedback alignment and diversity studies reveal limitations of demonstration-conditioned supervision~\citep{feedbackalign2026,sdsddiversity2026}. UniOPSD explicitly arbitrates between comparable outcome and hindsight step credits before token refinement.

\paragraph{Memory and self-evolving agents.}
ReAct couples reasoning with action; Reflexion and Voyager retain feedback and skills~\citep{react2023,reflexion2023,voyager2023}. Skill-SD supplies teacher-only skill context~\citep{skillsd2026}. Cognitive Scaffold, CoEvoKG, and SESA develop persistent graph memory or evolving task and skill stores~\citep{cognitivescaffold2026,coevokg2026,sesa2026}. UniOPSD uses peers within the current rollout group, making credit arbitration complementary to persistent memory.
% ===== BACKGROUND =====
\section{Background and Notation}
\label{sec:background}

For a task $x$, the behavior policy $\pi_b$ samples trajectories $\{\tau_i\}_{i\in\mathcal I_x}$ from the same initial environment state. Trajectory $i$ receives total return $G_i=\sum_{\ell=1}^{T_i}r_{i,\ell}$. We index an individual response by $k$, write its trajectory as $i(k)$, its interaction step as $\ell(k)$, its visible history as $h_k$, and its valid tokens as $y_{k,1:L_k}$. Its discounted return-to-go is $R_k=\sum_{\ell=\ell(k)}^{T_{i(k)}}\gamma^{\ell-\ell(k)}r_{i(k),\ell}$, where $\gamma\in(0,1]$. An anchor group $g(k)$ collects responses assigned to the same comparison state within a task, following GiGPO~\citep{gigpo2025}. The comparison state can be an observation representation rather than the agent's complete history, so matching an anchor does not itself assert identical latent states or continuation distributions.

GiGPO estimates relative advantages at two levels: complete trajectories are compared within a task, and actions are compared within an anchor group. Let $\mathcal R_x^E=(G_j)_{j\in\mathcal I_x}$ and $\mathcal R_k^S=(R_j)_{j\in g(k)}$ denote the corresponding return collections. Following its original formulation~\citep{gigpo2025},
\begin{align}
A^E(\tau_i)&=\frac{G_i-\operatorname{mean}(\mathcal R_x^E)}{F_{\mathrm{norm}}(\mathcal R_x^E)},
& A^S(a_k)&=\frac{R_k-\operatorname{mean}(\mathcal R_k^S)}{F_{\mathrm{norm}}(\mathcal R_k^S)},
\label{eq:outcome-branches}
\end{align}
where $a_k$ denotes the action response and $F_{\mathrm{norm}}$ is either the group standard deviation or the constant one. The episode term evaluates the trajectory as a whole and is shared across its responses; the step term distinguishes actions through their subsequent returns. GiGPO combines them as
\begin{equation}
A^{\mathrm{GiGPO}}(a_k)=A^E(\tau_{i(k)})+\omega A^S(a_k),
\label{eq:gigpo-advantage}
\end{equation}
where $\omega\geq0$ controls the step contribution. All reported runs use $F_{\mathrm{norm}}\equiv1$, the mean-centered variant. In the Method, $A_k^E$ denotes the episode contribution supplied by the backbone and $A_k^S$ denotes the weighted step contribution $\omega A^S(a_k)$. UniOPSD retains the former and arbitrates between the latter and hindsight credit. Appendix~\ref{app:gigpo-conventions} specifies the reported backbone's episode-baseline and reward conventions relative to the trajectory-level definition above.

Anchor centering defines a relative comparison. It does not supply missing counterfactual rollouts. In particular, a response can have $A_k^S=0$ even when other returns in its group vary, because its return equals the group mean. We therefore distinguish a zero-valued branch from a group without return variation. This distinction determines the availability masks used below and prevents a numerical zero from automatically being interpreted as missing environmental evidence.
% ===== METHOD =====
\section{Method}
\label{sec:method}

\begin{figure}[!htbp]
\centering
\includegraphics[width=\linewidth]{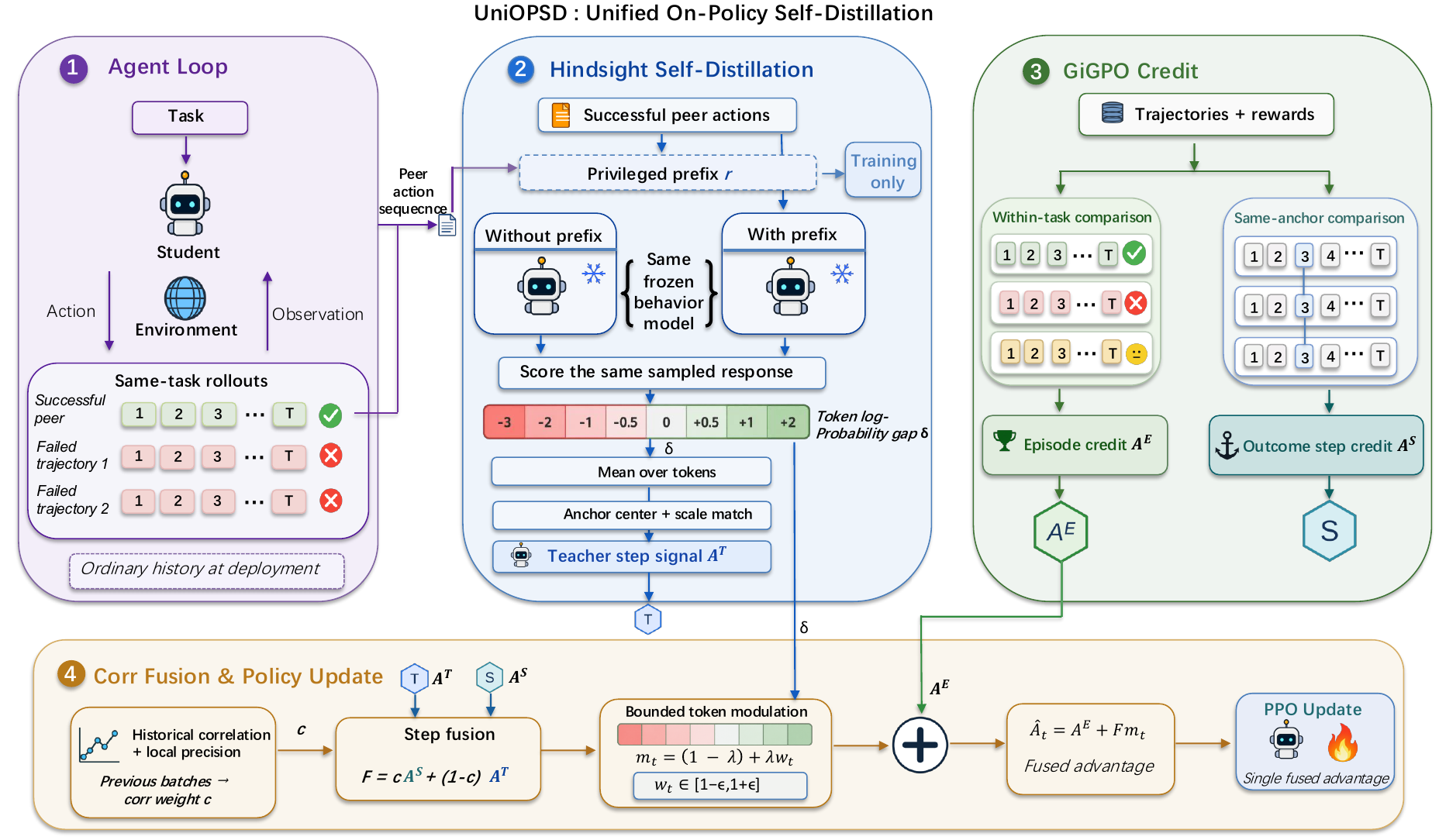}
\caption{UniOPSD arbitrates between outcome and hindsight step credit at matched anchors. Historical agreement sets a global mixing level, and local evidence adjusts each step's weight. The episode term is retained explicitly; bounded token modulation $b_{k,t}=1-\lambda_m+\lambda_m w_{k,t}$ refines the arbitrated step term before PPO. Privileged peer context is restricted to training.}
\label{fig:method}
\end{figure}

UniOPSD assigns influence to outcome and hindsight credit using evidence at two scales. Historical association sets the global mixing level, and current availability and relative precision determine the local allocation. The main variant implements this arbitration through the \texttt{corr} weighting rule and the \texttt{fusion} construction, retaining the backbone's episode contribution $A_k^E$ while replacing its weighted step contribution $A_k^S$. Its inputs are sampled trajectories, rewards, anchor assignments, and log probabilities from a frozen behavior checkpoint. No additional teacher parameters are fitted for the update. Figure~\ref{fig:method} summarizes the training pipeline.

\subsection{Comparable local credit estimates}
\label{sec:signals}

\paragraph{Hindsight signal construction.}
For each failed trajectory, we select the first successful peer in the same rollout task group and extract its ordered action sequence as a privileged prefix $r_k$. This follows the peer-hindsight construction used by StepOPSD~\citep{stepopsd2026}.
The teacher scores the student's already sampled response, rather than generating a replacement action. Both scores use the same frozen behavior checkpoint:
\begin{equation}
\delta_{k,t}=\log\pi_b(y_{k,t}\mid r_k,h_k,y_{k,<t})
-\log\pi_b(y_{k,t}\mid h_k,y_{k,<t}),
\qquad
\bar\delta_k=\frac{1}{L_k}\sum_{t=1}^{L_k}\delta_{k,t}.
\label{eq:gap}
\end{equation}
Successful trajectories and failed trajectories without a usable successful peer receive no prefix. Their teacher inputs coincide with the student inputs, yielding a zero gap. This policy makes teacher coverage a property of the sampled group, rather than a uniformly available supervision source.

To compare this hindsight feedback with native outcome credit, we first center the row-mean gaps over all rows in the corresponding anchor group, including rows with zero gaps. We then match the spread of the two step branches on the set of taught rows:
\begin{equation}
d_k=\bar\delta_k-\frac{1}{|g(k)|}\sum_{j\in g(k)}\bar\delta_j,
\qquad
A_k^T=s\,d_k,
\qquad
s=\frac{\operatorname{std}_{k\in\mathcal T}(A_k^S)}
{\operatorname{std}_{k\in\mathcal T}(d_k)+\varepsilon}.
\label{eq:teacher-branch}
\end{equation}
Here $\varepsilon>0$ is a numerical stability constant and $\mathcal T=\{k:\sum_t|\delta_{k,t}|>0\}$ is the set of taught rows. We set $s=0$ if fewer than two taught rows exist or their centered gaps have standard deviation at most $\varepsilon$. The scale population is $\mathcal T$, even when some taught anchors lack return variation. Common centering makes the comparisons consistent, and scale matching controls their numerical magnitude. These operations do not establish that the branches estimate the same true advantage.

\paragraph{Availability and relative precision.}
Local arbitration depends on whether each signal is available and how much its observations vary. Let $n_k=|g(k)|$, let $v_k^R$ be the sample variance of returns in that anchor, and let $v_k^\delta=L_k^{-1}\sum_t(\delta_{k,t}-\bar\delta_k)^2$. We define
\begin{equation}
\mathcal E=\{k:n_k\geq2,\ v_k^R>0\},\qquad
\mathcal B=\mathcal E\cap\mathcal T,
\qquad
p_k^S=\frac{n_k}{\max(v_k^R,\varepsilon)},\quad
p_k^T=\frac{1}{\max(s^2v_k^\delta/L_k,\varepsilon)}.
\label{eq:precision}
\end{equation}
The respective proxy is set to zero outside its availability set. Each positive proxy is divided by its own median over $\mathcal B$, producing $z_k^S$ and $z_k^T$. If $\mathcal B$ is empty, each median uses that source's own available rows. Empty populations use a unit denominator, and denominators are floored by $\varepsilon$. These are batch-relative precision proxies: small observed scatter is neither a measurement of bias nor a guarantee of useful supervision.

\subsection{Arbitration from historical and local evidence}
\label{sec:corr}

Local stability does not establish relevance to returns: a teacher can produce consistent credit without agreeing with outcome comparisons. Conversely, historical association cannot determine how much evidence supports each current step. UniOPSD therefore uses historical agreement to set a global mixing level and relative local precision to adjust the allocation on rows with both sources. On each valid batch, we compute Pearson correlation between $A^S$ and $A^T$ over $\mathcal B$, provided both vectors have nonzero variance. An exponential moving average with decay $\beta\in[0,1)$ stores these measurements. The current batch uses the previous stored value; its measurement updates the state only after the weights have been formed.

Let $\bar\rho_{m-1}$ denote the available historical average and $n_{\mathrm{prev}}$ the row count of its most recent valid measurement. We use
\begin{equation}
\widehat\rho_m=\max\left(0,\bar\rho_{m-1}-\frac{\kappa}{\sqrt{H\max(n_{\mathrm{prev}},1)}}\right),
\qquad
q_m=\min\{\widehat\rho_m^2,u(1-\eta)\},
\label{eq:rho}
\end{equation}
Here $\kappa\geq0$ controls shrinkage, $H=(1-\beta)^{-1}$ is a heuristic averaging horizon, and $\eta\in(0,1)$ is a numerical margin that keeps $q_m$ below $u$. We set $\widehat\rho_m=0$ before the first valid measurement. Because rows and batches share trajectories and changing policies, the subtracted quantity is not a calibrated confidence bound. The sensitivity parameter $u\in(0,1]$ is a design choice rather than measured reliability. Table~\ref{tab:uniopsd-settings} in Appendix~\ref{app:experiments} lists the parameter values used for credit-signal analysis.

The global outcome weight and the row-specific fusion weight are
\begin{align}
\bar c_m&=\frac{u(u-q_m)}{u^2-2u q_m+q_m},
\label{eq:cbar}\\
c_k&=
\begin{cases}
\displaystyle\frac{a_m z_k^S}{a_m z_k^S+z_k^T},& k\in\mathcal B,\\
\bar c_m,& k\in\mathcal T\setminus\mathcal E,\\
1,& k\notin\mathcal T,
\end{cases}
\qquad
a_m=\frac{\bar c_m}{1-\bar c_m},
\qquad
F_k=c_kA_k^S+(1-c_k)A_k^T.
\label{eq:fusion}
\end{align}
We set $\bar c_m=1$ when its denominator is nonpositive and clip it to $[0,1]$. If $\bar c_m\geq1-\eta$, all rows use $c_k=1$, avoiding the odds ratio. Remaining numerical denominators are floored by $\varepsilon$ and final weights clipped to $[0,1]$.

The coefficient $c_k$ expresses how much influence the outcome estimate receives, and $1-c_k$ is the hindsight share. These weights are assigned continuously, including when the estimates agree; the rule has no separate sign-conflict trigger. On teacher-only rows, $F_k=(1-\bar c_m)A_k^T$ because $A_k^S=0$. Rows without a teacher retain $A_k^S$, even if anchor centering gives their intermediate $d_k$ a nonzero value. A nonpositive shrunken correlation rejects the teacher step branch but does not remove token modulation. Historical agreement and relative precision are reliability proxies; the arbitration rule carries no optimal-estimation guarantee.

For jointly available branches, the interior rule has the odds form $c_k/(1-c_k)=[\bar c_m/(1-\bar c_m)](z_k^S/z_k^T)$. Equal normalized precision proxies retain the historical allocation, while a larger outcome-to-hindsight precision ratio increases the outcome share. When $A_k^S>0>A_k^T$, the fused credit is positive exactly when $c_k>-A_k^T/(A_k^S-A_k^T)$. The corresponding threshold therefore depends on the magnitudes of the conflicting estimates as well as their signs. This explains how one historical mixing level can support different decisions within a batch: local precision changes the allocation, and the two credit magnitudes determine the resulting direction.

\subsection{Bounded token modulation and policy update}
\label{sec:token}

After allocating step credit, we use the token gap for a bounded refinement within the sampled response:
\begin{align}
w_{k,t}&=\operatorname{clip}\!\left(\exp\{\operatorname{sign}(F_k)\delta_{k,t}\},1-\epsilon_w,1+\epsilon_w\right),
\label{eq:token-weight}\\
\widehat A_{k,t}&=A_k^E+F_k\bigl[(1-\lambda_m)+\lambda_m w_{k,t}\bigr],
\qquad
\lambda_m=\lambda_0\max\left(1-\frac{m}{T_{\mathrm{decay}}},0\right).
\label{eq:advantage}
\end{align}
Here $\lambda_0\in[0,1]$ is the initial modulation strength, $T_{\mathrm{decay}}>0$ is the decay duration in policy updates, and $\epsilon_w\in[0,1)$ is the clipping radius of the token weight. The multiplier depends on the direction of $F_k$ and acts only on its contribution. Its initial relative effect is bounded by $\lambda_0\epsilon_w$ and decays to zero over the schedule, while step arbitration remains active. The frozen gaps, group statistics, and resulting advantages are held fixed during the policy update.

With token ratio $r_{k,t}(\theta)=\pi_\theta(y_{k,t}\mid h_k,y_{k,<t})/\pi_b(y_{k,t}\mid h_k,y_{k,<t})$, the PPO surrogate is
\begin{equation}
J_{\mathrm{clip}}(\theta)=\mathbb E_{k,t}\!\left[
\min\!\left(r_{k,t}(\theta)\widehat A_{k,t},
\operatorname{clip}(r_{k,t}(\theta),1-\epsilon_{\mathrm{PPO}},1+\epsilon_{\mathrm{PPO}})\widehat A_{k,t}\right)\right].
\label{eq:ppo}
\end{equation}
Here $\epsilon_{\mathrm{PPO}}$ is the policy-ratio clipping radius, distinct from the token-weight radius $\epsilon_w$. This is the core clipped surrogate with the substituted advantage~\citep{ppo2017}. The implementation also uses a dual-clipping floor for negative advantages, detailed in Appendix~\ref{app:experiments}.
The training configuration retains its reference-policy KL control, entropy regularization, and invalid-action penalties. These inherited choices must be held fixed in controlled comparisons. At deployment, the policy uses ordinary interaction histories without peer hindsight.

\subsection{Properties and controlled reductions}
\label{sec:properties}

\paragraph{Proposition 1.}
For $c_k,\lambda_m\in[0,1]$ and $0\leq\epsilon_w<1$, the fused step value lies between $A_k^S$ and $A_k^T$, and
\begin{equation}
\left|\widehat A_{k,t}-(A_k^E+F_k)\right|\leq\lambda_m\epsilon_w|F_k|.
\label{eq:bound}
\end{equation}
Setting $c_k=1$ and $\lambda_m=0$ recovers $A_k^E+A_k^S$. The proof is in Appendix~\ref{app:properties}.

The reductions distinguish two interventions. Setting $c_k=1$ removes step fusion but retains token modulation, whereas setting $\lambda_m=0$ removes token modulation but retains step fusion. An untaught row has $c_k=1$ and $w_{k,t}=1$, so it recovers the underlying advantage directly. Keeping $A_k^E$ fixed places no conservation constraint on total trajectory credit and does not preserve the sign of the total advantage. These distinctions define interpretable controls for the experiments.
% ===== EXPERIMENTS =====
\section{Experiments}
\label{sec:experiments}

\subsection{Evaluation setting}
We evaluate UniOPSD on ALFWorld, WebShop, and Search-QA, covering household interaction, product selection, and retrieval-based question answering~\citep{alfworld2021,webshop2022,searchr12025}. The policies are Qwen2.5-3B-Instruct and Qwen2.5-7B-Instruct~\citep{qwen252024}. UniOPSD uses successful-peer trajectories from the same rollout group as training-time hindsight, without externally supplied skill descriptions. ALFWorld reports task success rates. WebShop reports both the mean task score, which includes partial credit, and the fraction of fully successful tasks. Search-QA reports accuracy on Natural Questions (NQ), TriviaQA (Triv), PopQA (Pop), HotpotQA (Hotp), 2Wiki (2Wk), MuSiQue (MuS), and Bamboogle (Bam).

For credit-signal analysis, we use correlation-guided peer fusion with eight rollouts per task and the mean-centered GiGPO backbone. The maximum interaction lengths are 50, 15, and 4 turns for ALFWorld, WebShop, and Search-QA. Token modulation initially changes the fused step term by at most ten percent and decays to zero over $T_{\mathrm{decay}}$ updates, while step arbitration continues. Search-QA diagnostics use the NQ/HotpotQA validation mixture, which differs from the seven-dataset benchmark comparison. Appendix~\ref{app:experiments} records the diagnostic settings, and Table~\ref{tab:uniopsd-settings} gives the shared method parameters.

% BEGIN MAIN RESULTS TABLE
% Transcribed from AgentOPSD arXiv:2608.05987v1, Table 1 (PDF page 6).
% Generated by scripts/import_agentopsd_table.py; AgentOPSD rows excluded.
\begin{table}[t]
\centering
\caption{Performance comparison of UniOPSD and baselines on ALFWorld, Search-QA, and WebShop. ALFWorld reports success rate, Search-QA reports accuracy, and WebShop reports score and accuracy (all in \%). $^{*}$ denotes validation with skills. Purple/bold and blue mark the highest and second-highest values, respectively, within each model and metric; ties share a color.}
\label{tab:reported-baselines}
\begingroup
\definecolor{SourceBest}{HTML}{E2D4F0}
\definecolor{SourceSecond}{HTML}{D9EAF7}
\definecolor{SourceModelBand}{HTML}{F2F2F2}
\scriptsize
\setlength{\tabcolsep}{2pt}
\renewcommand{\arraystretch}{1.13}
\resizebox{\linewidth}{!}{%
\begin{tabular}{@{}l@{\hspace{7pt}}*{17}{r}@{}}
\toprule
& \multicolumn{7}{c}{\textbf{ALFWorld}} & \multicolumn{8}{c}{\textbf{Search-QA}} & \multicolumn{2}{c}{\textbf{WebShop}} \\
\cmidrule(lr){2-8}\cmidrule(lr){9-16}\cmidrule(lr){17-18}
\textbf{Method} & \textbf{Pick} & \textbf{Look} & \textbf{Clean} & \textbf{Heat} & \textbf{Cool} & \textbf{Pick2} & \textbf{Avg} & \textbf{NQ} & \textbf{Triv} & \textbf{Pop} & \textbf{Hotp} & \textbf{2Wk} & \textbf{MuS} & \textbf{Bam} & \textbf{Avg} & \textbf{Score} & \textbf{Acc} \\
\midrule
\rowcolor{SourceModelBand}
\multicolumn{18}{l}{\textit{Qwen2.5-3B-Instruct}} \\
Vanilla & 44.4 & 11.1 & 6.2 & 15.4 & 28.6 & 12.5 & 21.9 & 24.6 & 48.1 & 31.0 & 26.3 & 25.3 & 7.2 & 59.7 & 31.7 & 6.7 & 0.8 \\
Skill-Prompt$^{*}$ & 51.7 & 66.7 & 48.4 & 0.0 & 4.3 & 10.0 & 28.9 & 23.7 & 46.2 & 30.6 & 24.4 & 22.1 & 7.5 & 12.5 & 23.9 & 0.2 & 0.8 \\
OPSD & 48.8 & 41.7 & 16.7 & 0.0 & 15.8 & 16.7 & 28.1 & 0.1 & 0.1 & 0.1 & 0.0 & 0.0 & 0.0 & 0.0 & 0.0 & 11.3 & 3.1 \\
GRPO & 91.2 & 62.5 & 96.2 & 61.9 & 65.0 & 47.4 & 75.0 & 39.3 & \cellcolor{SourceSecond}60.6 & 41.1 & 37.4 & 34.6 & 15.4 & 26.4 & 36.4 & 79.8 & 63.3 \\
Skill-GRPO & 88.9 & 71.4 & 58.8 & 70.6 & 40.7 & 29.2 & 60.2 & 43.5 & 58.8 & 43.0 & 36.8 & 32.2 & 11.7 & 12.5 & 34.1 & 77.3 & 60.9 \\
Skill-GRPO$^{*}$ & 94.3 & 57.1 & \cellcolor{SourceBest}\textbf{100} & 66.7 & 73.1 & 57.1 & 80.5 & 44.3 & 59.6 & 44.3 & 39.0 & 36.1 & 14.5 & 14.9 & 36.1 & 76.3 & 66.4 \\
GRPO+OPSD & \cellcolor{SourceBest}\textbf{100} & \cellcolor{SourceBest}\textbf{82.4} & 85.7 & \cellcolor{SourceSecond}75.0 & 70.0 & 60.0 & 81.2 & \cellcolor{SourceBest}\textbf{44.9} & \cellcolor{SourceBest}\textbf{61.2} & \cellcolor{SourceSecond}45.2 & \cellcolor{SourceBest}\textbf{40.4} & 38.5 & \cellcolor{SourceSecond}16.0 & \cellcolor{SourceSecond}66.1 & \cellcolor{SourceSecond}44.6 & 77.8 & 66.4 \\
Skill-SD & 88.2 & 50.0 & 96.2 & 52.4 & 65.0 & 57.9 & 73.4 & 44.4 & 60.4 & 44.0 & \cellcolor{SourceSecond}39.5 & \cellcolor{SourceBest}\textbf{40.4} & 15.4 & 64.9 & 44.1 & 75.9 & 64.0 \\
RLSD & 87.9 & \cellcolor{SourceSecond}75.0 & 90.9 & \cellcolor{SourceSecond}75.0 & 73.1 & 68.4 & 79.7 & 41.5 & 58.6 & 42.3 & \cellcolor{SourceBest}\textbf{40.4} & \cellcolor{SourceSecond}40.2 & \cellcolor{SourceBest}\textbf{16.8} & \cellcolor{SourceBest}\textbf{66.9} & 43.8 & 84.4 & 66.4 \\
SDAR & \cellcolor{SourceSecond}97.1 & 62.5 & \cellcolor{SourceBest}\textbf{100} & 61.9 & \cellcolor{SourceBest}\textbf{75.0} & \cellcolor{SourceBest}\textbf{84.2} & \cellcolor{SourceBest}\textbf{84.4} & \cellcolor{SourceSecond}44.8 & 58.1 & 44.3 & 38.6 & 36.2 & 15.7 & \cellcolor{SourceSecond}66.1 & 43.4 & \cellcolor{SourceSecond}85.0 & \cellcolor{SourceSecond}68.0 \\
StepOPSD & 82.4 & 66.7 & 82.6 & 52.2 & \cellcolor{SourceSecond}73.7 & 75.0 & 73.4 & 43.6 & \cellcolor{SourceBest}\textbf{61.2} & 43.8 & 39.2 & 38.1 & 15.8 & 64.5 & 43.7 & 82.4 & 66.4 \\
% Local evaluation provenance: data/uniopsd_alfworld_step240_seed256.json.
% Source variant: invvar; evaluated update 240; evaluation seed 256; 128 episodes.
% Local evaluation provenance: data/uniopsd_webshop_3b.json.
% Source variant: corr; evaluated update 150.
UniOPSD & 94.7 & 64.3 & \cellcolor{SourceSecond}96.3 & \cellcolor{SourceBest}\textbf{100.0} & 44.4 & \cellcolor{SourceSecond}77.8 & \cellcolor{SourceSecond}82.8 & 43.8 & \cellcolor{SourceBest}\textbf{61.2} & \cellcolor{SourceBest}\textbf{46.0} & 39.0 & 39.8 & 14.9 & \cellcolor{SourceSecond}66.1 & \cellcolor{SourceBest}\textbf{45.3} & \cellcolor{SourceBest}\textbf{87.4} & \cellcolor{SourceBest}\textbf{75.0} \\
\midrule
\rowcolor{SourceModelBand}
\multicolumn{18}{l}{\textit{Qwen2.5-7B-Instruct}} \\
Vanilla & 36.1 & 22.2 & 3.1 & 0.0 & 0.0 & 0.0 & 12.5 & 25.2 & 50.8 & 29.5 & 29.0 & 29.0 & 10.4 & 63.7 & 33.9 & 5.9 & 1.6 \\
Skill-Prompt$^{*}$ & 51.7 & 50.0 & 32.3 & 5.3 & 4.3 & 0.0 & 23.4 & 30.9 & 52.1 & 32.7 & 32.7 & 27.9 & 12.7 & 66.1 & 36.4 & 1.7 & 0.8 \\
OPSD & 50.0 & 60.0 & 22.7 & 21.4 & 17.6 & 9.5 & 32.8 & 8.8 & 8.6 & 17.5 & 2.5 & 4.2 & 0.5 & 1.2 & 6.2 & 4.5 & 2.3 \\
GRPO & 91.2 & 87.5 & 96.2 & 81.0 & 65.0 & 57.9 & 81.2 & 45.1 & 63.7 & 44.0 & 43.6 & 43.2 & 16.8 & 37.6 & 42.0 & 80.9 & 72.6 \\
Skill-GRPO & 88.5 & 66.7 & 65.2 & 61.1 & 57.7 & \cellcolor{SourceSecond}73.1 & 69.5 & 45.2 & 63.7 & 45.7 & 43.1 & 43.3 & 19.6 & 21.4 & 40.3 & 80.4 & 71.9 \\
Skill-GRPO$^{*}$ & \cellcolor{SourceBest}\textbf{100} & 83.3 & \cellcolor{SourceSecond}96.4 & 83.3 & 75.0 & \cellcolor{SourceBest}\textbf{78.9} & \cellcolor{SourceSecond}88.3 & 44.8 & 63.0 & 45.1 & 43.7 & 43.7 & \cellcolor{SourceSecond}20.5 & 71.4 & 47.5 & 87.0 & 81.2 \\
GRPO+OPSD & 91.4 & 61.5 & \cellcolor{SourceBest}\textbf{100} & 87.5 & \cellcolor{SourceSecond}76.5 & 52.2 & 80.4 & \cellcolor{SourceSecond}47.3 & 64.5 & 46.9 & 43.8 & 39.3 & 18.0 & 69.4 & 47.0 & 86.8 & 76.5 \\
Skill-SD & 93.9 & \cellcolor{SourceBest}\textbf{93.8} & 90.9 & \cellcolor{SourceBest}\textbf{100} & 69.2 & 68.4 & 85.1 & 47.1 & 64.5 & 47.8 & 44.2 & 42.1 & 20.2 & 69.0 & 47.8 & 86.1 & 76.5 \\
RLSD & \cellcolor{SourceBest}\textbf{100} & 87.5 & 92.3 & 58.8 & \cellcolor{SourceBest}\textbf{80.0} & 65.2 & 82.0 & 46.8 & 63.0 & 44.4 & \cellcolor{SourceBest}\textbf{45.5} & \cellcolor{SourceBest}\textbf{48.9} & \cellcolor{SourceBest}\textbf{21.5} & \cellcolor{SourceBest}\textbf{73.0} & \cellcolor{SourceSecond}49.0 & 87.4 & 77.3 \\
SDAR & 94.7 & 75.0 & \cellcolor{SourceBest}\textbf{100} & 86.7 & 68.2 & \cellcolor{SourceBest}\textbf{78.9} & 85.9 & 46.3 & 63.5 & \cellcolor{SourceSecond}48.2 & 43.8 & \cellcolor{SourceSecond}48.4 & 19.6 & \cellcolor{SourceBest}\textbf{73.0} & \cellcolor{SourceSecond}49.0 & \cellcolor{SourceBest}\textbf{89.4} & \cellcolor{SourceBest}\textbf{82.8} \\
StepOPSD & \cellcolor{SourceSecond}98.1 & 75.0 & \cellcolor{SourceBest}\textbf{100} & \cellcolor{SourceSecond}90.5 & \cellcolor{SourceBest}\textbf{80.0} & 63.2 & \cellcolor{SourceBest}\textbf{88.4} & 45.3 & \cellcolor{SourceSecond}64.6 & 45.1 & 44.5 & 44.4 & 19.3 & 69.8 & 48.2 & 87.2 & 78.1 \\
% Local evaluation provenance: data/uniopsd_alfworld_7b_step130.json.
% Source variant: invvar; evaluated update 130.
% Local evaluation provenance: data/uniopsd_webshop_7b.json.
% Source variant: invvar; evaluated update 150.
UniOPSD & 89.7 & \cellcolor{SourceSecond}90.0 & 94.7 & 88.2 & 66.7 & 72.0 & 83.6 & \cellcolor{SourceBest}\textbf{48.2} & \cellcolor{SourceBest}\textbf{64.8} & \cellcolor{SourceBest}\textbf{49.1} & \cellcolor{SourceSecond}44.7 & 46.0 & 20.0 & \cellcolor{SourceSecond}72.8 & \cellcolor{SourceBest}\textbf{49.8} & \cellcolor{SourceSecond}88.8 & \cellcolor{SourceSecond}82.0 \\
\bottomrule
\end{tabular}%
}
\endgroup
\end{table}
% END MAIN RESULTS TABLE

\subsection{Main results}
UniOPSD achieves ALFWorld success rates of $82.8\%$ and $83.6\%$ with the 3B and 7B models, respectively. The 3B result ranks second among the methods in Table~\ref{tab:reported-baselines}, behind SDAR's $84.4\%$; the 7B result is below StepOPSD's $88.4\%$. On WebShop, the 3B model reaches a score of $87.4$ and success rate of $75.0\%$, exceeding SDAR's $85.0$ and $68.0\%$ by $2.4$ score points and $7.0$ percentage points. Both metrics are the highest in the 3B comparison. The 7B model reaches $88.8$ and $82.0\%$, ranking second on both metrics, within $0.6$ score points and $0.8$ percentage points of SDAR. Thus the largest aggregate improvement over the reported baselines occurs in 3B WebShop, while performance on ALFWorld remains below the strongest baseline at each scale.

Search-QA shows complementary strengths across the constituent datasets. The 3B model attains the highest PopQA accuracy of $46.0\%$ and ties the highest TriviaQA accuracy at $61.2\%$. The 7B model leads on NQ, TriviaQA, and PopQA with $48.2\%$, $64.8\%$, and $49.1\%$, respectively, and ranks second on HotpotQA and Bamboogle. The reported aggregate accuracies are $45.3\%$ and $49.8\%$. Without externally supplied skill descriptions, UniOPSD outperforms most compared methods on the aggregate metrics reported in Table~\ref{tab:reported-baselines}. On WebShop, the 3B model achieves a success rate of $75.0\%$, exceeding the strongest reported baseline by $7.0$ percentage points. Table~\ref{tab:backbone-ablation} compares UniOPSD with outcome-only backbones and simplified fusion rules.
% Source Appendix E defines Search-QA Avg as pooled accuracy over all evaluation questions.
% Validate aggregates from subset sample counts or logs, not an assumed equal-weight mean.

\Needspace{6\baselineskip}
\subsection{Component ablations}
Table~\ref{tab:backbone-ablation} compares five configurations that examine outcome credit, hindsight supervision, and the fusion rule. Removing the teacher sets $c_k\equiv1$ and $\lambda_m\equiv0$, recovering GiGPO and removing both the hindsight step branch and token modulation. GRPO provides a trajectory-only backbone reference by additionally removing anchor-relative step credit from this outcome-only learner. Replacing correlation-guided fusion with inverse-variance weighting (\texttt{invvar}) retains the two credit branches and local precision adjustment, but removes the historical correlation prior. The frozen-correlation variant retains the correlation-to-weight mapping and local adjustment while fixing the correlation input to $\rho_0$ throughout training, with $\rho_0=0.25$. This distinguishes a fixed correlation input from an evolving historical signal. Appendix~\ref{app:controls} gives the ablation definitions.

UniOPSD has the highest aggregate result in all six model--benchmark settings in Table~\ref{tab:backbone-ablation}. Its gains over GiGPO range from $3.9$ to $7.0$ percentage points. Relative to inverse-variance fusion, UniOPSD is higher by $3.9$, $4.0$, and $3.1$ points on ALFWorld, Search-QA, and WebShop with the 3B model, and by $5.5$, $3.5$, and $5.4$ points with the 7B model. The frozen-correlation variant falls between inverse-variance fusion and UniOPSD in every setting, with UniOPSD ahead by $1.6$--$3.1$ points. For example, 7B WebShop success is $76.6\%$ with inverse-variance fusion, $78.9\%$ with a fixed correlation input, and $82.0\%$ for UniOPSD.

% BEGIN BACKBONE ABLATION TABLE
% GRPO values are copied from Table 1 as backbone reference values, not a local matched rerun.
% GiGPO aggregates are author-entered values synchronized from the manuscript repository.
% UniOPSD retains the author-selected Table 1 reference values: 82.8, 45.3, 75.0, 83.6, 49.8, 82.0.
% This mixed-variant reference row is not a configuration-matched corr evaluation.
% UniOPSD 3B ALFWorld is the Table 1 invvar update-240 evaluation, not Figure 3's corr update 125.
% UniOPSD 3B WebShop: data/uniopsd_webshop_3b.json, corr update 150.
% UniOPSD 3B ALFWorld (invvar): data/uniopsd_alfworld_step240_seed256.json, update 240.
% UniOPSD 7B ALFWorld (invvar): data/uniopsd_alfworld_7b_step130.json, update 130.
% UniOPSD 7B WebShop (invvar): data/uniopsd_webshop_7b.json, update 150.
% Search-QA aggregates 45.3/49.8 are author-supplied reference values; per-variant provenance is pending.
% Completed invvar and frozen-rho aggregates are author-entered results synchronized from
% manuscript commit 8e68c08; see data/uniopsd_ablation_results.json.
% Exact training logs and matched-protocol verification for these new rows are not archived here.
\begin{table}[!t]
\centering
\caption{Component ablations using overall benchmark results (\%). ALFWorld and WebShop report success rates; Search-QA reports aggregate accuracy. The UniOPSD row reproduces Table~\ref{tab:reported-baselines} for reference; the frozen variant fixes $\rho_0=0.25$.}
\label{tab:backbone-ablation}
\begingroup
\small
\setlength{\tabcolsep}{3.5pt}
\renewcommand{\arraystretch}{1.15}
\resizebox{\linewidth}{!}{%
\begin{tabular}{@{}lrrrrrr@{}}
\toprule
& \multicolumn{3}{c}{\textbf{Qwen2.5-3B-Instruct}} & \multicolumn{3}{c}{\textbf{Qwen2.5-7B-Instruct}} \\
\cmidrule(lr){2-4}\cmidrule(lr){5-7}
\textbf{Variant} & ALFWorld & Search-QA & WebShop & ALFWorld & Search-QA & WebShop \\
\midrule
w/o step grouping (GRPO) & 75.0 & 36.4 & 63.3 & 81.2 & 42.0 & 72.6 \\
w/o teacher (GiGPO) & 76.6 & 39.8 & 70.3 & 79.7 & 45.6 & 75.0 \\
w/o correlation (\texttt{invvar}) & 78.9 & 41.3 & 71.9 & 78.1 & 46.3 & 76.6 \\
w/ frozen $\rho$ & 80.5 & 42.8 & 73.4 & 82.0 & 47.5 & 78.9 \\
\textbf{UniOPSD} & 82.8 & 45.3 & 75.0 & 83.6 & 49.8 & 82.0 \\
\bottomrule
\end{tabular}%
}
\endgroup
\end{table}
% END BACKBONE ABLATION TABLE

\begin{figure}[!t]
\centering
\includegraphics[width=\linewidth]{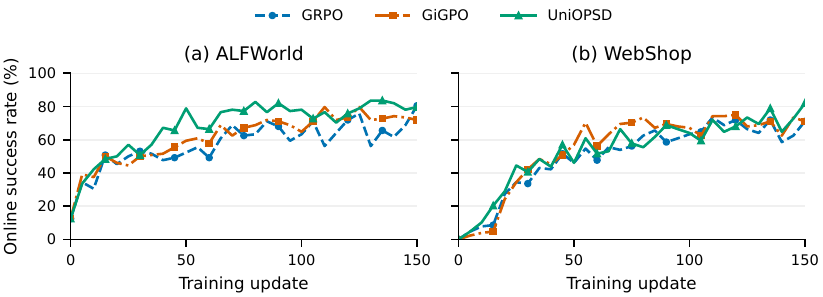}
\caption{Online validation success rates of GRPO, GiGPO, and UniOPSD with Qwen2.5-7B-Instruct on ALFWorld and WebShop, measured every five training updates. UniOPSD uses successful-peer hindsight with correlation-guided fusion (\texttt{corr}).}
\label{fig:ablation-7b}
\end{figure}

Figure~\ref{fig:ablation-7b} complements the aggregate comparison with 7B training curves on ALFWorld and WebShop. On WebShop, UniOPSD reaches $82.0\%$ success at update 150, while GRPO and GiGPO both finish at $71.1\%$. On ALFWorld, UniOPSD reaches $83.6\%$ during training and finishes at $79.7\%$, compared with $80.5\%$ for GRPO and $71.9\%$ for GiGPO at the final update.

\subsection{Training dynamics}
\begin{figure}[!t]
\centering
\includegraphics[width=\linewidth]{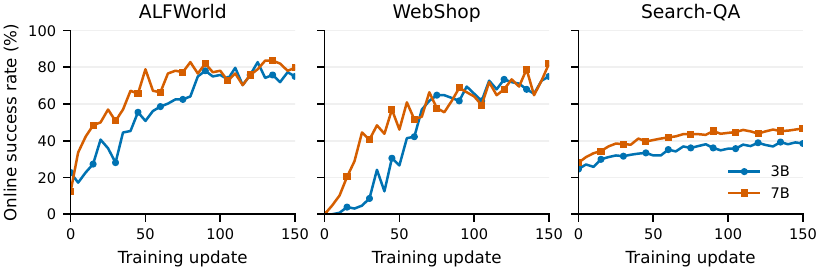}
\caption{UniOPSD validation success rates for Qwen2.5-3B/7B-Instruct. ALFWorld and WebShop 7B curves match Figure~\ref{fig:ablation-7b} and use correlation-guided fusion (\texttt{corr}).}
\label{fig:pilot}
\end{figure}

Figure~\ref{fig:pilot} presents the learning curves of Qwen2.5-3B-Instruct and Qwen2.5-7B-Instruct across the three environments over 150 training updates. We evaluate the policy every five updates. ALFWorld and WebShop use 128 validation samples with sampling temperature $0.4$, while Search-QA uses 512 samples with deterministic decoding. ALFWorld uses the in-distribution evaluation split.

Both model scales improve across the three environments, with earlier gains on ALFWorld and WebShop and more gradual improvement on Search-QA. The 3B ALFWorld curve reaches $82.8\%$ at update 125 and ends at $75.0\%$ at update 150, illustrating that peak performance and the final checkpoint need not coincide.

Over 150 retained ALFWorld-3B updates, UniOPSD uses $5.7\%$ fewer student tokens and $4.7\%$ less training time per update than our local SDAR reproduction (accounting exclusions: Appendix~\ref{app:runtime-cost}).

\FloatBarrier
\subsection{Local disagreement and evidence coverage}
We measure agreement and evidence coverage throughout training. Over updates 1--150, mean within-batch correlation on rows carrying both signals is $0.315$ for ALFWorld-3B and $0.308$ for ALFWorld-7B, while sign-disagreement fractions are $37.8\%$ and $36.8\%$. Disagreement includes zero versus nonzero signs as well as opposing nonzero signs. Both signals are available on $32.8\%$ and $27.0\%$ of rows on average. Thus, positive average association coexists with substantial local disagreement and incomplete coverage. Fusion also assigns nonzero step credit to some responses whose outcome-step credit is zero: the fraction satisfying $A_k^S=0$ and $F_k\ne0$ averages $5.6\%$ and $4.7\%$. Appendix~\ref{app:diagnostics} gives definitions and results for all domains.

WebShop further separates coverage from conditional disagreement: both signals are available on $10.1\%$ and $12.9\%$ of rows for the 3B and 7B models, yet disagree on $26.9\%$ and $30.0\%$ of those rows. The all-row mean outcome weights, $0.976$ and $0.974$, include rows without teacher feedback, where the outcome weight is one. These averages therefore do not imply negligible hindsight influence wherever it is available. Coverage and conditional agreement together clarify where arbitration can change the update.

% ===== CONCLUSION =====
\section{Conclusion}
\label{sec:conclusion}

UniOPSD unifies outcome and hindsight supervision through adaptive local credit arbitration. Historical agreement sets the global mixing level, while current availability and relative precision determine each decision's allocation before bounded token refinement. Experiments with 3B and 7B policies show competitive performance across ALFWorld, WebShop, and Search-QA. Component comparisons favor UniOPSD over inverse-variance and frozen-correlation fusion, while diagnostics reveal positive average agreement alongside substantial local disagreement. Together, these results support conditioning the influence of hindsight on both historical and current evidence.

\clearpage
\section*{AI Assistance Disclosure}
Generative AI tools were used solely to assist with language polishing and to improve the clarity, grammar, and readability of author-written text. They were not used to generate research ideas, design the methodology or experiments, derive mathematical claims, or interpret experimental results. All AI-assisted edits were carefully reviewed and verified by the authors, who take full responsibility for the final content of the paper.

\section*{Reproducibility Statement}
Appendix~\ref{app:method} details the algorithm, implementation conventions, and proofs. Appendix~\ref{app:experiments} documents training and evaluation settings, including the use of eight NVIDIA A100 GPUs, and the provenance of reported measurements. Appendix~\ref{app:diagnostics} defines the diagnostic statistics, while Appendix~\ref{app:prompts-examples} provides interaction prompts and qualitative examples. The code and configurations required to reproduce our experiments are available at \url{https://github.com/Zenghuang-Fu/Uniopsd}.

\bibliography{iclr2027_conference}
\bibliographystyle{iclr2027_conference}

\clearpage
\appendix
% Keep appendix float-only pages top-aligned as well.
\makeatletter
\setlength{\@fptop}{0pt}
\makeatother
% ===== APPENDIX METHOD =====
\section{Method Specification and Additional Analysis}
\label{app:method}

\subsection{Training procedure}
\label{app:algorithm}

Table~\ref{alg:uniopsd} specifies one training iteration. All comparisons use the current rollout batch, whereas the global correlation state is inherited from earlier valid batches. A batch with fewer than two jointly available rows, or zero variance in either comparison vector, supplies no valid correlation measurement. In that case, the stored average and its associated row count remain unchanged. The first valid measurement initializes the average directly. Later valid measurements update it as $\bar\rho\leftarrow\beta\bar\rho+(1-\beta)\rho$, using the EMA decay defined in Section~\ref{sec:corr}. The row count is that of the latest valid batch, not a cumulative sample count.

\begin{table}[!t]
\caption{One UniOPSD iteration. All teacher scoring and advantage construction are performed without gradients. $\bar\rho$ and $n_{\mathrm{prev}}$ denote the stored historical state.}
\label{alg:uniopsd}
\centering
\begin{tabular}{p{0.035\linewidth}p{0.90\linewidth}}
\hline
1 & Freeze the behavior checkpoint $\pi_b$ and sample trajectory groups for the current tasks. Record trajectory returns, discounted step returns, anchor assignments, response masks, and student log probabilities.\\[3pt]
2 & Select a successful peer for each task group. Give its extracted action sequence to failed trajectories only. Score the same sampled tokens with $\pi_b$ under this prefix and compute Eq.~\eqref{eq:gap}.\\[3pt]
3 & Construct $A_k^E$ and the weighted $A_k^S$ using Appendix~\ref{app:gigpo-conventions}. Compute $d_k$ over every row of each anchor and the scale $s$ over $\mathcal T$. Form $A^T$ using Eq.~\eqref{eq:teacher-branch}.\\[3pt]
4 & Construct $\mathcal E$, $\mathcal T$, and $\mathcal B$. Compute the precision proxies and their separate median normalizations, with source-specific fallback populations.\\[3pt]
5 & Read the stored correlation state. Compute $\widehat\rho_m$, $\bar c_m$, and $c_k$, including the all-outcome endpoint. Form the fused step value $F_k$.\\[3pt]
6 & If the current $\mathcal B$ yields a valid Pearson correlation, update $\bar\rho$ and $n_{\mathrm{prev}}$ for the next iteration. This update does not change the already formed $c_k$.\\[3pt]
7 & Compute the scheduled token weights and $\widehat A_{k,t}$ from Eqs.~\eqref{eq:token-weight}--\eqref{eq:advantage}, retaining the response mask.\\[3pt]
8 & Optimize the clipped policy surrogate with the configured reward penalties and regularization. Keep the stored log probabilities, gaps, and advantages fixed during this update.\\
\hline
\end{tabular}
\end{table}

\subsection{Backbone normalization and episode-baseline conventions}
\label{app:gigpo-conventions}

All six reported runs set \texttt{algorithm.gigpo.mode=mean\_norm}, which selects $F_{\mathrm{norm}}\equiv1$ for both GiGPO branches, and use $\omega=1$. The alternative \texttt{mean\_std\_norm} divides the centered values by their group standard deviations and is not used in these runs. The coefficient $\omega$ is absorbed into $A_k^S$ throughout the Method, including teacher scale matching, so it must not be applied a second time in the fusion formula.

The trajectory-level episode definition in Eq.~\eqref{eq:outcome-branches} compares each trajectory once. The inherited implementation instead computes its default episode baseline across flattened response rows within each task. Let $\mathcal J_x$ contain these rows and let $e_k$ be the sum of token rewards supplied to the episode branch for row $k$. The actual backbone contributions used by the reported runs are
\begin{equation}
A_k^E=e_k-b_x,\qquad
b_x=\begin{cases}
\displaystyle\frac{1}{|\mathcal J_x|}\sum_{j\in\mathcal J_x}e_j,&|\mathcal J_x|\geq2,\\
0,&|\mathcal J_x|=1,
\end{cases}
\qquad
A_k^S=\omega\left(R_k-\frac{1}{|g(k)|}\sum_{j\in g(k)}R_j\right).
\label{eq:implemented-outcome}
\end{equation}
Here $R_k$ is the discounted return-to-go after any configured row-level invalid-action penalty. Both scalar contributions are repeated over the response's valid tokens. If $e_k=G_{i(k)}$, a trajectory represented by $M_i$ rows contributes $M_i$ copies of its return to $b_x$, giving $b_x=\sum_{i\in\mathcal I_x}M_iG_i/\sum_{i\in\mathcal I_x}M_i$ for a nonsingleton task group. Thus this baseline can differ from the equal-trajectory mean even though both use $F_{\mathrm{norm}}\equiv1$. Row-level penalties can also make $e_k$ vary within a trajectory. UniOPSD leaves this supplied episode contribution unchanged; the algebraic properties in Section~\ref{sec:properties} hold for either baseline convention. Controlled comparisons must retain the same sampling, reward, and baseline conventions.

\subsection{Masking, populations, and numerical conventions}
\label{app:masks}

The three populations in the method have different purposes. $\mathcal E$ measures whether an anchor offers at least two rows with nonzero return variance. $\mathcal T$ detects a nonzero token gap, so a supplied prefix that leaves every token score unchanged does not create teacher availability. Their intersection $\mathcal B$ is the population on which both proxies are normalized and their correlation is measured. In contrast, scale matching uses all of $\mathcal T$. Replacing that population by $\mathcal B$ changes the algorithm whenever taught rows lack return variation.

Peer extraction uses action-bearing spans in trajectory order and excludes the peer's full intermediate reasoning. Where responses use search and answer tags, those tagged spans supply the corresponding action text. Repeated copies of a trajectory step created by batch padding are removed from the peer action list. A peer without extractable actions yields no usable prefix. Prefix limits and prompt truncation determine what the teacher actually sees. Together with the extracted peer actions and sampled response tokens, they specify the information used to form the gap before fusion.

The anchor mean for $d_k$ includes untaught rows. Therefore, an untaught row with $\bar\delta_k=0$ can have $d_k\neq0$ if other rows in its anchor have nonzero gaps. Its availability mask still forces $c_k=1$, preventing this centering effect from creating a teacher contribution on that row. This also means that the effective teacher contribution $(1-c_k)A_k^T$ need not sum to zero within an anchor, even though the unweighted $d_k$ does. Neither anchor centering nor retention of the episode term implies conservation of total credit.

For $n_k\geq2$, anchor return variance uses the sample denominator $n_k-1$. Token-gap variance instead uses the population denominator $L_k$ over valid response tokens. The standard deviations in Eq.~\eqref{eq:teacher-branch} use sample conventions. Response padding is excluded from each token reduction. A singleton anchor has zero return variance and zero centered teacher gap. If the scale cannot be estimated, setting $s=0$ makes every teacher step value zero, but teacher availability remains determined by the original gap. In particular, availability refers to the presence of the input signal, not a proof that the processed branch is informative.

A zero scale does not automatically restore the outcome-only step branch. If a nonzero gap keeps a row in $\mathcal T$ and stored correlation gives $c_k<1$, then $A_k^T=0$ yields $F_k=c_kA_k^S$. The rule can therefore attenuate a nonzero outcome step value without contributing nonzero teacher step credit.

The median normalizations divide each source by a typical value in the same jointly available population. This avoids directly comparing quantities with incompatible raw units. When that population is empty, the environmental denominator uses $\mathcal E$ and the teacher denominator uses $\mathcal T$. If a source is entirely absent, its normalized proxy remains zero. The medians are computed over rows, as in the implementation, so an anchor with more rows can contribute repeatedly to the environmental proxy distribution. Group-weighted normalization would be a distinct variant.

\subsection{Interpretation of the correlation rule}
\label{app:correlation}

The correlation rule imposes a particular response to observed association. At zero shrunken correlation, $q_m=0$ and $\bar c_m=1$. Step fusion then selects the outcome branch everywhere. The nonnegative truncation treats negative association as a reason to reject the teacher step contribution, rather than reverse its sign. With $u=0.5$, Eq.~\eqref{eq:cbar} simplifies to $\bar c_m=1-2q_m$ before endpoint handling. Increasing positive association consequently reduces the nominal outcome weight in this setting. Neither this monotonic behavior nor the squared-correlation form estimates causal usefulness.

Within $\mathcal B$, the proxy ratio changes the row weight around this global setting. If $z_k^S=z_k^T$, the resulting $c_k$ equals $\bar c_m$. Separate median normalization does not require these two normalized values to coincide on any particular row, and it does not guarantee that the batch median of $c_k$ equals $\bar c_m$. On teacher-only rows, the method applies the global shrinkage directly because there is no nonzero environmental precision with which to arbitrate. The same global decision can therefore have different effects on the jointly available and teacher-only populations.

The uncertainty subtraction in Eq.~\eqref{eq:rho} is deliberately described as heuristic. The averaging horizon $H$ does not count independent replications, and the most recent row count does not measure the effective sample size of the average. Rows share trajectories, peers, and often anchors, while successive policies and task mixtures can change the correlation. The current correlation measurement must not be mistaken for the lagged value used to construct the current weights.

\subsection{Proof of the stated properties}
\label{app:properties}

For any two real values $A_k^S,A_k^T$ and $c_k\in[0,1]$, their convex combination satisfies
\begin{equation}
\min(A_k^S,A_k^T)\leq F_k\leq\max(A_k^S,A_k^T).
\end{equation}
The clipped token weight obeys $|w_{k,t}-1|\leq\epsilon_w$. Subtracting the unmodulated advantage gives
\begin{equation}
\widehat A_{k,t}-(A_k^E+F_k)=\lambda_mF_k(w_{k,t}-1).
\end{equation}
Taking absolute values proves Eq.~\eqref{eq:bound}. Setting $c_k=1$ gives $F_k=A_k^S$, and additionally setting $\lambda_m=0$ removes the multiplier, proving recovery of $A_k^E+A_k^S$. These statements concern the constructed advantage. They require no distributional assumptions and imply no equivalence between the optimized policy and an outcome-only optimum.

The distinction between branch direction and total advantage matters even under the perturbation bound. For example, let $A_k^E=0.95$, $F_k=-1$, $\lambda_m=0.5$, and $w_{k,t}=0.8$. The unmodulated total is $-0.05$, while the modulated total is $0.05$. Thus a bounded multiplier can reverse the total advantage near cancellation, even though its positive factor leaves the direction of the fused step contribution unchanged. Similarly, step fusion can change the total advantage whenever the two branches differ. No total-sign guarantee follows from Proposition~1.

\subsection{Ablation configurations}
\label{app:controls}

The outcome-only control in Table~\ref{tab:backbone-ablation} sets both $c_k\equiv1$ and $\lambda_m\equiv0$, recovering GiGPO. Its episode and anchor-relative step terms remain active, while the hindsight branch and token modulation are removed. GRPO retains only the trajectory-level outcome comparison. The inverse-variance variant retains hindsight credit and token modulation, but removes the historical correlation prior. With unit prior and no extra gate, it uses $c_k=z_k^S/(z_k^S+z_k^T)$ when either source is available, with the deterministic all-absent fallback. These definitions distinguish removal of the teacher from a change in how the two local estimates are combined.

The frozen-correlation control replaces the effective correlation input $\widehat\rho_m$ in Eq.~\eqref{eq:rho} by $\rho_0$ at every update, with $\rho_0=0.25$. It retains the same $u$, the mapping in Eq.~\eqref{eq:cbar}, local precision adjustment, availability masks, and token schedule. Thus the global correlation input is fixed, while the row weights $c_k$ can still vary with local evidence.

\begin{figure}[!t]
\centering
\includegraphics[width=.68\linewidth]{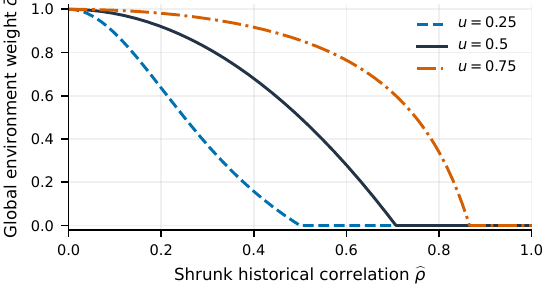}
\caption{Analytic response of the global correlation controller, not empirical performance. The main method uses $u=0.5$; alternative settings illustrate its sensitivity. The squared correlation is capped below $u$ as specified in the algorithm. Row-level precision ratios further modify these global weights.}
\label{fig:controller}
\end{figure}
% ===== APPENDIX EXPERIMENTS =====
\section{Experimental details and provenance}
\label{app:experiments}

\paragraph{Implementation settings.}
Training was conducted on eight NVIDIA A100 GPUs.
ALFWorld and WebShop use exact anchor grouping within each task; Search-QA uses the backbone's similarity grouping with threshold $0.9$. For credit-signal analysis, we use $F_{\mathrm{norm}}\equiv1$ (\texttt{mean\_norm}) and $\omega=1$, with the episode-baseline convention specified in Appendix~\ref{app:gigpo-conventions}. The shared UniOPSD parameters are in Table~\ref{tab:uniopsd-settings}. The actor learning rate is $10^{-6}$, with one PPO epoch per update and PPO clipping radius $\epsilon_{\mathrm{PPO}}=0.2$. Rollout temperature is $1.0$. ALFWorld and WebShop validation uses sampling at temperature $0.4$; Search-QA validation disables sampling with temperature zero. Training starts from the corresponding instruction-tuned Qwen2.5 model. The peer teacher is a conditional branch of the behavior policy; it is not a separately trained checkpoint. The training objective also retains the implementation's reference-policy KL, entropy, and invalid-action terms. These regularizers are distinct from a privileged-teacher auxiliary distillation loss.

\begin{table}[!t]
\centering
\caption{UniOPSD parameter settings for credit-signal analysis. The horizon $H$ is derived from the EMA decay; the numerical margin $\eta$ and stability constant $\varepsilon$ serve different roles despite having the same value.}
\label{tab:uniopsd-settings}
\begin{tabular}{llr}
\toprule
Parameter & Symbol & Value \\
\midrule
Global sensitivity & $u$ & $0.5$ \\
Correlation EMA decay & $\beta$ & $0.9$ \\
Correlation shrinkage strength & $\kappa$ & $2.33$ \\
Derived averaging horizon & $H=(1-\beta)^{-1}$ & $10$ \\
Initial token modulation & $\lambda_0$ & $0.5$ \\
Token modulation decay duration & $T_{\mathrm{decay}}$ & $100$ updates \\
Token-weight clipping radius & $\epsilon_w$ & $0.2$ \\
Numerical stability constant & $\varepsilon$ & $10^{-6}$ \\
Numerical endpoint margin & $\eta$ & $10^{-6}$ \\
\bottomrule
\end{tabular}
\end{table}

\begin{table}[!t]
\centering
\caption{Training settings for credit-signal analysis at both model scales within each environment. Prompt and response lengths are token limits per model call. The PPO minibatch size counts flattened training rows rather than complete episodes.}
\label{tab:hyperparameters}
\begin{tabular}{lrrr}
\toprule
Setting & ALFWorld & WebShop & Search-QA \\
\midrule
Training tasks per batch & 16 & 16 & 128 \\
Rollouts per task & 8 & 8 & 8 \\
Validation batch size & 128 & 128 & 512 \\
Maximum interaction turns & 50 & 15 & 4 \\
History length & 2 & 2 & 4 \\
Maximum prompt tokens & 2048 & 4096 & 4096 \\
Maximum response tokens & 512 & 512 & 512 \\
PPO minibatch size & 256 & 64 & 256 \\
Reference KL coefficient & 0.01 & 0.01 & 0.001 \\
Entropy coefficient & 0.001 & 0.001 & 0.001 \\
Invalid-action coefficient & 0.1 & 0.1 & 0.01 \\
Validation interval (updates) & 5 & 5 & 5 \\
Validation temperature & 0.4 & 0.4 & 0 \\
Similarity-based anchors & No & No & Yes ($0.9$) \\
\bottomrule
\end{tabular}
\end{table}

\paragraph{Loss aggregation and dual clipping.}
The actor aggregates valid-token losses by their mean. If $\ell_{k,t}$ denotes the minimum inside Eq.~\eqref{eq:ppo}, the configured implementation replaces it by $\max(\ell_{k,t},3\widehat A_{k,t})$ when $\widehat A_{k,t}<0$, and leaves it unchanged otherwise. This dual-clipping floor, together with the regularizers above, is inherited from the training backbone and should be matched across ablations.

\paragraph{Task and teacher information.}
ALFWorld uses the configured in-distribution split and WebShop uses the implementation's small-catalog setting. Search-QA uses the configured Natural Questions and HotpotQA validation mixture with top-three retrieval and at most four interaction turns. The successful peer's action sequence is retained for interactive environments. For Search-QA, both search queries and the successful final answer are present in the training teacher context. The student rollout and validation policy receive neither this hindsight prefix nor the peer answer. Nevertheless, answer-containing teacher context is a stronger training information source than search queries alone.

\paragraph{Evaluation and aggregation.}
Learning curves display raw online validation measurements at five-update intervals. Credit-signal statistics average valid batch-level measurements over updates 1--150, using the populations defined in Appendix~\ref{app:diagnostics}. Benchmark aggregates in Table~\ref{tab:reported-baselines} and the online validation curves summarize their respective evaluation records. Since the source values are rounded console outputs, the reported rates and diagnostic means should not be interpreted as higher-precision episode-level estimates.

\paragraph{Runtime comparison with SDAR.}
\label{app:runtime-cost}
Table~\ref{tab:runtime-cost} compares the ALFWorld-3B UniOPSD run in Figure~\ref{fig:pilot} with our local reproduction of SDAR~\citep{sdar2026}. Both use Qwen2.5-3B-Instruct, eight NVIDIA A100-SXM4-80GB GPUs, 16 tasks per batch, eight rollouts per task, and environment seed zero. Model, data, environment, learning rate, minibatch size, and parallelism settings match. SDAR retains its GRPO backbone and gated distillation loss, while UniOPSD uses GiGPO with correlation-guided peer fusion.

Statistics cover the 150 updates retained in each training path. SDAR combines original updates 1--45 with updates 46--150 after resuming from checkpoint 45. The discarded original updates 46--59 and time outside the recorded updates are excluded. Within each update, we subtract validation and checkpoint-saving time from the logged iteration time.

\begin{table}[!t]
\centering
\caption{Runtime comparison with a local SDAR reproduction on ALFWorld-3B over 150 retained updates. Student tokens total the prompt and response tokens in the training batches. Training time excludes validation and checkpoint saving. Reductions use SDAR as the reference.}
\label{tab:runtime-cost}
\begin{tabular}{lrrr}
\toprule
Method & \shortstack{Student tokens\\(M, cumulative)} & \shortstack{Teacher-stage time\\(s/update)} & \shortstack{Training time\\(s/update)} \\
\midrule
SDAR & 432.61 & 59.22 & 427.63 \\
UniOPSD & 407.76 & 22.11 & 407.43 \\
\midrule
Reduction (\%) & 5.7 & 62.7 & 4.7 \\
\bottomrule
\end{tabular}
\end{table}

The teacher stage includes prefix construction, tokenization, and likelihood scoring. Student token totals include repeated history and rows copied for batch divisibility, but exclude privileged teacher prefixes and repeated forward or backward passes. Each method has one historical training path, executed on different dates. The comparison describes observed costs of the two methods; it does not isolate the effect of the teacher source or establish convergence efficiency at matched success rates across training seeds.

\FloatBarrier
\section{Definitions of the diagnostic measurements}
\label{app:diagnostics}

Let $\mathcal B_m$ contain rows with a teacher gap and a nondegenerate anchor return group at update $m$. The logged step correlation is the Pearson correlation of $A^S$ and $A^T$ restricted to $\mathcal B_m$, when both vectors have nonzero variance. Sign disagreement is the fraction of these rows with $\operatorname{sign}(A_k^S)\ne\operatorname{sign}(A_k^T)$. Table~\ref{tab:diagnostics} averages each logged statistic over updates 1--150; when a statistic is undefined at an update it is omitted rather than set to zero. These are means of batch statistics, not a pooled correlation or a pooled episode-level frequency.

\begin{table}[!t]
\centering
\caption{Step-signal statistics across model scales and environments over updates 1--150. Availability is the fraction of rows in $\mathcal B_m$; disagreement is conditional on that set. The final column is the all-row mean environment mixing weight and consequently includes rows without a teacher.}
\label{tab:diagnostics}
\begin{tabular}{llrrrr}
\toprule
Environment & Model & Correlation & Disagreement (\%) & Availability (\%) & Mean $c_k$ \\
\midrule
ALFWorld & 3B & 0.315 & 37.8 & 32.8 & 0.897 \\
ALFWorld & 7B & 0.308 & 36.8 & 27.0 & 0.928 \\
WebShop & 3B & 0.302 & 26.9 & 10.1 & 0.976 \\
WebShop & 7B & 0.272 & 30.0 & 12.9 & 0.974 \\
Search-QA & 3B & 0.371 & 26.8 & 11.7 & 0.950 \\
Search-QA & 7B & 0.446 & 20.6 & 9.9 & 0.949 \\
\bottomrule
\end{tabular}
\end{table}

The log also measures the fraction of all rows satisfying $A_k^S=0$ and $F_k\ne0$. This counts algebraic activation of a previously zero step term. It neither measures additional correct actions nor proves improvement in return estimation. Similarly, a large correlation can reflect shared rollout selection and common outcomes. Source-level association, evidence availability, and downstream learning benefit must be measured separately.

\clearpage
\section{Prompt templates and qualitative examples}
\label{app:prompts-examples}

We provide the interaction prompts used by the agent and the privileged prefix used by the training teacher, followed by selected successful validation excerpts. The interaction templates come from the shared agent backbone. Braced fields are populated by the environment; line wrapping and typography are normalized for readability. The examples show observable decisions without introducing additional performance measures.

\subsection{Student interaction prompts}
\label{app:interaction-prompts}

The following are the history-bearing templates. At the first interaction, the corresponding initial template omits the history; ALFWorld's initial observation also contains the task description. The student sees its own interaction context and the available environment actions.

\begin{PromptPanel}{ALFWorld Interaction Prompt}
You are an expert agent operating in the ALFRED Embodied Environment. Your task is to: \PromptField{task\_description}
\par
Prior to this step, you have already taken \PromptField{step\_count} step(s). Below are the most recent \PromptField{history\_length} observations and the corresponding actions you took: \PromptField{action\_history}
\par
You are now at step \PromptField{current\_step} and your current observation is: \PromptField{current\_observation}
\par
Your admissible actions of the current situation are: [\PromptField{admissible\_actions}].
\par\smallskip
Now it's your turn to take an action.
\par
You should first reason step-by-step about the current situation. This reasoning process MUST be enclosed within \CaseTag{CaseThink}{think} \CaseTag{CaseThink}{/think} tags.
\par
Once you've finished your reasoning, you should choose an admissible action for current step and present it within \CaseTag{CaseAction}{action} \CaseTag{CaseAction}{/action} tags.
\end{PromptPanel}

\begin{PromptPanel}{WebShop Interaction Prompt}
You are an expert autonomous agent operating in the WebShop e-commerce environment.
\par
Your task is to: \PromptField{task\_description}.
\par
Prior to this step, you have already taken \PromptField{step\_count} step(s). Below are the most recent \PromptField{history\_length} observations and the corresponding actions you took: \PromptField{action\_history}
\par
You are now at step \PromptField{current\_step} and your current observation is: \PromptField{current\_observation}.
\par
Your admissible actions of the current situation are:
\par
[\PromptField{available\_actions}].
\par\smallskip
Now it's your turn to take one action for the current step.
\par
You should first reason step-by-step about the current situation, then think carefully which admissible action best advances the shopping goal. This reasoning process MUST be enclosed within \CaseTag{CaseThink}{think} \CaseTag{CaseThink}{/think} tags.
\par
Once you've finished your reasoning, you should choose an admissible action for current step and present it within \CaseTag{CaseAction}{action} \CaseTag{CaseAction}{/action} tags.
\end{PromptPanel}

\clearpage
\begin{PromptPanel}{Search-QA Interaction Prompt}
You are an expert agent tasked with answering the given question step-by-step.
\par
Your question: \PromptField{task\_description}
\par\smallskip
Prior to this step, you have already taken \PromptField{step\_count} step(s). Below is the interaction history where \CaseTag{CaseAction}{search} \CaseTag{CaseAction}{/search} wrapped your past search queries and \CaseTag{CaseInfo}{information} \CaseTag{CaseInfo}{/information} wrapped the corresponding search results returned by the external search engine. History:
\par
\PromptField{memory\_context}
\par\smallskip
Now it's your turn to respond for the current step.
\par
You should first conduct reasoning process. This process MUST be enclosed within \CaseTag{CaseThink}{think} \CaseTag{CaseThink}{/think} tags.
\par
After completing your reasoning, choose only one of the following actions (do not perform both):
\par
(1) If you find you lack some knowledge, you can call a search engine to get more external information using format: \CaseTag{CaseAction}{search} your query \CaseTag{CaseAction}{/search}.
\par
(2) If you have enough knowledge to answer the question confidently, provide your final answer within \CaseTag{CaseAnswer}{answer} \CaseTag{CaseAnswer}{/answer} tags, without detailed illustrations. For example, \CaseTag{CaseAnswer}{answer}Beijing\CaseTag{CaseAnswer}{/answer}.
\end{PromptPanel}

\subsection{Privileged teacher context}
\label{app:teacher-prompt}

During training, UniOPSD selects the first successful peer from the same task group and constructs the prefix below for failed trajectories. The numbered entries are placeholders for the peer's extracted actions, in interaction order. ALFWorld and WebShop retain the command text inside each \texttt{<action>} span; Search-QA retains the \texttt{<search>} or \texttt{<answer>} span, including the successful final answer. Peer reasoning and retrieved document contents are not copied into this prefix.
\begin{PromptPanel}{Privileged Hindsight Prefix}
[Privileged Hindsight Information]
\par
Another attempt at this exact task succeeded. Its action sequence was:
\par
1. \PromptField{first\_peer\_action}
\par
2. \PromptField{second\_peer\_action}
\par
...
\par
\PromptField{n}. \PromptField{last\_peer\_action}
\end{PromptPanel}

The teacher input concatenates this prefix with the student's original serialized prompt and scores the same sampled response. Both likelihood evaluations use the frozen behavior checkpoint. Successful trajectories and task groups without a usable successful peer keep the original student input. The prefix is restricted to teacher scoring during training; rollout and validation use the student interaction prompts above, without peer hindsight or an external skill bank. Thus, the additional context changes the evidence available for evaluating a decision, while the sampled decision itself remains fixed.

\clearpage
\subsection{Selected successful examples}
\label{app:qualitative-cases}

The following transcripts show selected successful Qwen2.5-3B-Instruct UniOPSD validation episodes. Each presents the task, retained interaction history, and final response. Shaded passages reproduce the model's logged final-step reasoning; colored tags distinguish reasoning, actions, retrieved information, and answers. Observations are shortened with ellipses, and earlier actions are shown only where retained in the log.

\begin{table}[!t]
\centering
\caption{ALFWorld: cleaning an object before placing it at the target.}
\label{tab:case-alfworld}
\begin{PromptPanel}{ALFWorld Successful Interaction}
\textbf{Task:} \textit{put a clean lettuce in countertop.} \\
\hline
\textbf{Earlier action:} \texttt{clean lettuce 1 with sinkbasin 1} \\
\textbf{Observation:} You clean the lettuce 1 using the sinkbasin 1. \\
\textbf{Earlier action:} \texttt{go to countertop 1} \\
\textbf{Observation:} You arrive at countertop 1. \ldots \\
\rowcolor{CaseShade}\CaseTag{CaseThink}{think}Given the current situation, I have already cleaned the lettuce 1 using the sinkbasin 1. The next step should be to move the cleaned lettuce 1 to the countertop 1. The countertop 1 contains various items, but the lettuce 1 is not currently there. I need to move the lettuce 1 to the countertop 1 to complete the task.\CaseTag{CaseThink}{/think} \\
\CaseTag{CaseAction}{action}move lettuce 1 to countertop 1\CaseTag{CaseAction}{/action}
\end{PromptPanel}
\end{table}

\begin{table}[!t]
\centering
\caption{WebShop: selecting the requested color and size before purchase.}
\label{tab:case-webshop}
\begin{PromptPanel}{WebShop Successful Interaction}
\textbf{Task:} \textit{Find me machine wash men's t-shirts with long sleeve with color: swamp, and size: 3x-large big tall, and price lower than 70.00 dollars.} \\
\hline
\textbf{Earlier action:} \texttt{click[swamp]} \\
\textbf{Earlier action:} \texttt{click[3x-large big tall]} \\
\textbf{Observation:} \ldots Legendary Whitetails Men's Non-Typical Long Sleeve T-Shirt \ldots Price: \$10.52 to \$40.5 \ldots Buy Now. \\
\rowcolor{CaseShade}\CaseTag{CaseThink}{think}Based on the current observation, the search results show men's t-shirts with the color 'swamp' and size '3x-large big tall'. The price range is \$10.52 to \$40.50. The task is to find a t-shirt within this price range. The current action of clicking on 'swamp' and '3x-large big tall' is the first step. The next logical action would be to click on 'buy now' to purchase the t-shirt. I will now click on 'buy now' to initiate the purchase process.\CaseTag{CaseThink}{/think} \\
\CaseTag{CaseAction}{action}click[buy now]\CaseTag{CaseAction}{/action}
\end{PromptPanel}
\end{table}

\begin{table}[!t]
\centering
\caption{Search-QA: retrieving evidence about both entities before answering.}
\label{tab:case-search}
\begin{PromptPanel}{Search-QA Successful Interaction}
\textbf{Question:} \textit{Vocelli Pizza and Noble Roman's serve which kind of food?} \\
\hline
\CaseTag{CaseAction}{search}Vocelli Pizza and Noble Roman's serve which kind of food\CaseTag{CaseAction}{/search} \\
\CaseTag{CaseInfo}{information} \textbf{Doc 1 (Title: ``Noble Roman's'')} Noble Roman's is a pizza company based in Indianapolis, Indiana. \ldots\CaseTag{CaseInfo}{/information} \\
\CaseTag{CaseAction}{search}Vocelli Pizza food type\CaseTag{CaseAction}{/search} \\
\CaseTag{CaseInfo}{information} \textbf{Doc 1 (Title: ``Vocelli Pizza'')} Vocelli Pizza (formerly Pizza Outlet) is a pizzeria based in Pittsburgh, Pennsylvania. \ldots\CaseTag{CaseInfo}{/information} \\
\rowcolor{CaseShade}\CaseTag{CaseThink}{think}From the information provided, we know that Noble Roman's is a pizza company. Similarly, Vocelli Pizza is also described as a pizzeria. Therefore, both of these establishments serve pizza.\CaseTag{CaseThink}{/think} \\
\CaseTag{CaseAnswer}{answer}pizza\CaseTag{CaseAnswer}{/answer}
\end{PromptPanel}
\end{table}
\end{document}